\documentclass[sigconf]{acmart}

\usepackage{booktabs} 

\setcopyright{rightsretained}
\usepackage{multirow}
\usepackage{makecell}
\usepackage{tabularx}
\definecolor{candypink}{rgb}{0.89, 0.44, 0.48}

\acmDOI{10.475/3627631.3627669}

\acmISBN{123-4567-24-567/08/06}

\acmConference[ICVGIP'23]{14th Indian Conference on Computer Vision, Graphics and Image Processing}{December 2023}{Ropar, India}
\acmYear{2023}
\copyrightyear{2023}

\acmPrice{15.00}

\begin{document}
\title{\texttt{C-SAW}: Self-Supervised Prompt Learning for Image Generalization in Remote Sensing}

\author{Avigyan Bhattacharya}
 \affiliation{
   \institution{Indian Institute of Technology Bombay}
   \streetaddress{Powai}
   \city{Mumbai}
   \state{Maharashtra}
   \country{India}
   \postcode{400076}
 }
\author{Mainak Singha}
 \affiliation{
   \institution{Indian Institute of Technology Bombay}
   \streetaddress{Powai}
   \city{Mumbai}
   \state{Maharashtra}
   \country{India}
   \postcode{400076}
 }
\author{Ankit Jha}
 \affiliation{
   \institution{Indian Institute of Technology Bombay}
   \streetaddress{Powai}
   \city{Mumbai}
   \state{Maharashtra}
   \country{India}
   \postcode{400076}
 }
 \author{Biplab Banerjee}
 \affiliation{
   \institution{Indian Institute of Technology Bombay}
   \streetaddress{Powai}
   \city{Mumbai}
   \state{Maharashtra}
   \country{India}
   \postcode{400076}
 }

\renewcommand{\shortauthors}{}

\begin{abstract}
We focus on domain and class generalization problems in analyzing optical remote sensing images, using the large-scale pre-trained vision-language model (VLM), CLIP. While contrastively trained VLMs show impressive zero-shot generalization performance, their effectiveness is limited when dealing with diverse domains during training and testing. Existing prompt learning techniques overlook the importance of incorporating domain and content information into the prompts, which results in a drop in performance while dealing with such multi-domain data.
To address these challenges, we propose a solution that ensures domain-invariant prompt learning while enhancing the expressiveness of visual features. We observe that CLIP's vision encoder struggles to identify contextual image information, particularly when image patches are jumbled up. This issue is especially severe in optical remote sensing images, where land-cover classes exhibit well-defined contextual appearances.
To this end, we introduce \texttt{C-SAW}, a method that complements CLIP with a self-supervised loss in the visual space and a novel prompt learning technique that emphasizes both visual domain and content-specific features. We keep the CLIP backbone frozen and introduce a small set of projectors for both the CLIP encoders to train \texttt{C-SAW} contrastively. Experimental results demonstrate the superiority of \texttt{C-SAW} across multiple remote sensing benchmarks and different generalization tasks.
\end{abstract}

%
%



\maketitle

\section{Introduction}
\begin{figure}[htbp!]
    \centering
    \includegraphics[width = 7cm]{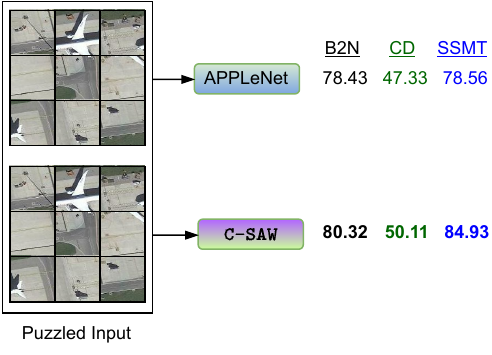}
    \vspace{-4.0mm}
    \caption{\texttt{C-SAW} shows better performance scores in comparison to APPLeNet \cite{applenet} for jumbled RS images. Here, B2N, CD and SSMT represent base-to-new class generalization, cross-dataset generalization and single source multi target generalization, respectively.}
    \label{fig:teaser}
    \vspace{-0.5cm}
\end{figure}

In recent times, the field of remote sensing (RS) imaging has witnessed remarkable advancements, revolutionizing Earth's surface monitoring for diverse applications, such as urban planning \cite{hino2018machine} and environmental monitoring \cite{sabins1999remote}. Representation learning, a common approach in this domain, involves pre-training a model on a large image dataset, like ImageNet \cite{krizhevsky2012imagenet}, using a supervised approach, showcasing significant improvements in various downstream tasks. However, the effectiveness of traditional deep learning models in analyzing complex RS images and outperforming ad-hoc machine learning methods is hampered by challenges in generalization when faced with domain shifts.

To tackle these challenges, researchers have explored the areas of domain adaptation (DA) \cite{ganin2015unsupervised, tuia2016domain}, which entails capturing representations of the target domain during model adaptation, drawing from knowledge obtained from the target distribution during training, and domain generalization (DG) \cite{li2018learning, rsdg1} addresses the practical scenario where the inclusion of RS data covering vast diversity during model training becomes arduous. DG capitalizes on labeled data from multiple source domains to learn a versatile and universal representation, striving to build an accurate model that adeptly handles any ``unseen" target domain. Despite the resounding success of DG in the computer vision field, its potential application in the context of RS imagery remains relatively unexplored.

Large multimodal foundation models, such as CLIP \cite{clip}, ALIGN \cite{align}, and ViLBERT \cite{lu2019vilbert}, have demonstrated impressive performance in downstream tasks, even with limited training data under zero-shot or few-shot learning scenarios. These models establish connections between image and text pairs through contrastive learning and effectively fine-tuning with hand-crafted text prompts. This success has paved the way for two exciting research directions. The first direction focuses on adapting pre-trained VLMs to diverse downstream tasks thus venturing into the area of transfer learning. Popular transfer approaches, such as prompt tuning \cite{coop, cocoop, maple, applenet} and visual adaptation \cite{clip-adapter, zhang2021tip}, also strive to achieve the desired objectives. On the other hand, the second direction explores knowledge distillation \cite{gu2021open, ding2022decoupling}, enhancing VLMs' performance in downstream tasks like object detection and semantic segmentation. Remote sensing images often face adverse conditions, like high altitudes or cloudiness, making it challenging to classify them using pre-trained CNN models. To address this, APPLeNet \cite{applenet} leverages prompt learning to generalize across remote sensing domains. Additionally, it employs context redundancy penalizing (CRP) loss to reduce redundancy between context tokens.

However, this paper dives into the promising area of self supervised learning, a technique that has garnered popularity due to its accomplishments in language and vision \cite{albert,cert,ssl_cv1,ssl_cv2}. Self-supervised learning offers an attractive alternate to supervised pre-training, guiding models to learn better embedding spaces. It has the potential to significantly enhance representation learning in RS image analysis, bolstering the generalization capabilities of models across diverse environments. By leveraging self-supervised learning and contrastive methods \cite{contrastive1, contrastive2}, our work emphasizes the prospects of advancing RS image analysis through domain generalization techniques, contributing to the domain's evolution and yielding valuable insights into RS applications. In order to incorporate the SSL task, we create the patches of the input image and then jumble it before passing through the pre-trained vision encoder to get the contextual latent embeddings.

As CLIP \cite{clip} is trained on massive datasets of image-text pairs, it is able to learn the relationship between images and their corresponding text descriptions, thus gaining recognition for its remarkable prowess. \textit{However, despite its success, CLIP encounters a  challenge in that it struggles to discern the positional relationships among different parts of an image, leading to occasional difficulties in processing jumbled imagery.} Let us assume a jumbled image where various parts have been intentionally scrambled, presenting a captivating puzzle for CLIP to solve. In such intriguing scenarios, VLMs are not able to rely solely on its pre-trained knowledge to identify distinct part embeddings of the image. Unfortunately, acquiring sufficient data for fine-tuning a VLM to conquer this challenge is not always feasible.

In remote sensing data for computer vision, dealing with scrambled satellite images poses a unique challenge. These images have parts out of order due to practical factors like data transmission errors or limitations in satellite imaging. Using pretrained CLIP directly on such jumbled images is difficult as the extracted features lack meaningful information. APPLeNet \cite{applenet} has achieved significant performance gains over CLIP and other learning methods for domain generalization in remote sensing images. However, APPLeNet encounters difficulties when dealing with jumbled RS images. As shown in Figure \ref{fig:teaser}, our demonstrations indicate that APPLeNet's performance decreases by approximately 0.2\% - 0.8\% on average across three types of domain generalization tasks (Base-to-New, Cross Dataset, and Single Source Multi Target) when presented with jumbled RS images instead of non-jumbled ones. To address this challenge of handling jumbled RS data effectively, we propose a method called \texttt{C-SAW}. It leverages a contrastive self-supervised learning framework for robust domain generalization. Through a context-aware self-supervision mechanism, we divide an image into smaller patches and rearrange them randomly. With SSL training involving reconstruction and self-supervised loss, our approach reconstructs the original image while learning robust contextual representations from the jigsaw inputs. These learned representations enable the model to efficiently perform diverse downstream tasks, including classification. In summary, our contributions are as follows:\\
\noindent - We propose a novel self-supervised prompt learning technique called \texttt{C-SAW} for image generalization in remote sensing applications.\\
\noindent - Our proposed method, \texttt{C-SAW}, tackles the limitations of part embeddings in CLIP by incorporating a reconstruction task to enhance the latent visual space of the distorted input image. Furthermore, we generate the visual attention tokens using $\mathcal{G_{VAT}}$ before the frozen text encoder in CLIP to impose desired prompt constraints on the input visual embeddings.\\
\noindent  - We extensively tested our approach on five optical remote sensing (RS) image classification benchmarks, evaluating its performance on three challenging generalization tasks: cross-dataset, base-to-new class, and single-source multi-target. Our results demonstrate that \texttt{C-SAW} surpasses the relevant literature by a significant margin, achieving a mean classification score improvement of approximately 2-6\%.

\section{Related Works}
\subsection{Self-Supervised Learning in Remote Sensing}
Stojnic et al. in \cite{stojnic2018evaluation, stojnic2021self} apply split-brain autoencoders \cite{splitbrain} and Contrastive Multiview Coding (CMC) \cite{cmc} on aerial images to learn effective representations for classification tasks while SatMAE \cite{cong2022satmae} uses a masked autoencoder modified for remote sensing data. Recent works in semantic segmentation include \cite{li2022global} with a global style-local matching contrastive learning approach and \cite{marsocci2021mare} for the Vaihingen dataset. FALSE \cite{zhang2022false} proposes efficient negative sampling, where as \cite{li2021geographical, ayush2021geography} addresses supervision noise and temporal alignment. Other contrastive approaches include \cite{tao2020remote}, where they use SSL approach to obtain high performance pre-training and \cite{kang2020deep}, where the authors make use of the semantic similarities among nearby scenes. \cite{vincenzi2021color} creates diverse samples through spatial and spectral transformations. GAN discriminators are used in \cite{dong2020self} and \cite{chen2021self} for temporal and multiview images, respectively. For detailed information, one can refer to \cite{tao2023self,wang2022self}.

\subsection{Domain Generalization}
One of the crucial tasks for deep learning models facing domain shift challenges between training and test distributions, which is mainly referred to as \textit{domain generalization} task. This task encompasses mainly two variants: multi-source domain generalization (Multi-DG) and single-source domain generalization (Single-DG). 
In Multi-DG, researchers have explored \textit{meta learning} approaches \cite{li2018learning,finn2017model} and subsequent works have built upon this, incorporating it for regularizers, semantic consistency, and feature critic losses \cite{5,6,li2019feature}. \textit{Adversarial training} methods \cite{li2018deep, matsuura2020domain, li2018domain} and \textit{Domain augmentation} techniques \cite{zhou2020deep, 8, 7, zhou2021domain, xu2021fourier} have been adopted to align feature distributions across different domains and generate new domains through adversarial examples. Other methods address Multi-DG through domain-specific masks, gradient-based dropout, episodic training, and style bias reduction techniques \cite{chattopadhyay2020learning, huang2020self, li2019episodic, nam2021reducing}.
Over the years, Single-DG has gained more attention as a practical and challenging problem. Domain expansion is a prevalent approach in this regard, \cite{volpi2018generalizing, qiao2021uncertainty, li2021progressive, qiao2020learning, 18, 26, xu2023simde} although different methods have also been proposed, including adversarial attacks with semantic restrictions, information bottleneck techniques, Wasserstein auto-encoders, contrastive learning, uncertainty estimation, and objective functions for domain expansion \cite{volpi2018generalizing, szegedy2013intriguing, 26, qiao2020learning, tishby2000information, tolstikhin2017wasserstein, li2021progressive, qiao2021uncertainty, xu2023simde}.
Despite these efforts, domain generalization for remote sensing (RS) image classification remains an area with limited attention to date \cite{rsdg1, rsdg2}.

\vspace{0.5cm}
\subsection{Vision-Language Models and Prompt Learning}
Large-scale Vision-Language Models (VLMs) fuse visual and textual inputs, enhancing performance in various computer vision tasks. Multimodal learning excels over unimodal approaches in feature learning, benefiting tasks like visual question answering (VQA) \cite{vqa}, image captioning \cite{shottell}, and image retrieval \cite{retrieval}, which rely on joint visual-semantic supervision. VLMs typically employ pre-trained language models like BERT \cite{bert} and GPT \cite{gpt} for textual encoding, while Convolutional Networks (ConvNets) or Vision Transformers process visual inputs. Notable VLMs include CLIP \cite{clip} and VisualBERT \cite{visualbert}.

Prompt learning gained popularity in NLP \cite{nlp} and visual recognition tasks, utilizing pre-trained language models like BERT to offer valuable information through textual prompts for downstream tasks. Automated prompt generation, explored in AutoPrompt \cite{shin2020autoprompt}, identifies tokens with significant gradient changes for prompt optimization. CoOp \cite{coop} fine-tunes CLIP for few-shot image classification, optimizing prompts, where as other methods, like ProGrad \cite{prograd}, generate prompts from visual features and optimize context tokens. Additionally, PDL \cite{pdl} proposes optimizing multiple prompt sets.

In the domain generalization for remote sensing images, APPLeNet \cite{applenet} introduces a vision-language prompting technique. SLIP \cite{slip} improves CLIP's performance by supplementing contrastive learning with a self-supervised learning objective in a multi-task setup. These methods showcase the growing interest in prompt learning to enhance language models' capabilities in visual recognition tasks.

CoCoOp \cite{cocoop} focuses on learning text prompts conditioned to input image embeddings, facilitating more context-aware queries. ProGrad \cite{prograd} employs prompt tuning through distillation, transferring knowledge from learned few-shot prompts to zero-shot prompts, which can be especially valuable for improving generalization. MaPLe \cite{maple} specializes in fine-tuning the CLIP model by aligning vision and text modalities, enhancing the model's overall performance. APPLeNet \cite{applenet} leverages attention mechanisms on conditioned image embeddings to improve text-prompting, simultaneously minimizing redundancy between context vectors for more efficient text conditioning on images. In contrast, \texttt{C-SAW} introduces a novel approach, that showcases the effectiveness of CLIP over \textit{image part embeddings by incorporating a reconstruction task (SSL)} to enhance the latent visual space of distorted input images. Additionally, it generates visual attention tokens to impose desired prompt constraints on the input visual embeddings.

\begin{figure*}
    \centering
    \includegraphics[width = \textwidth]{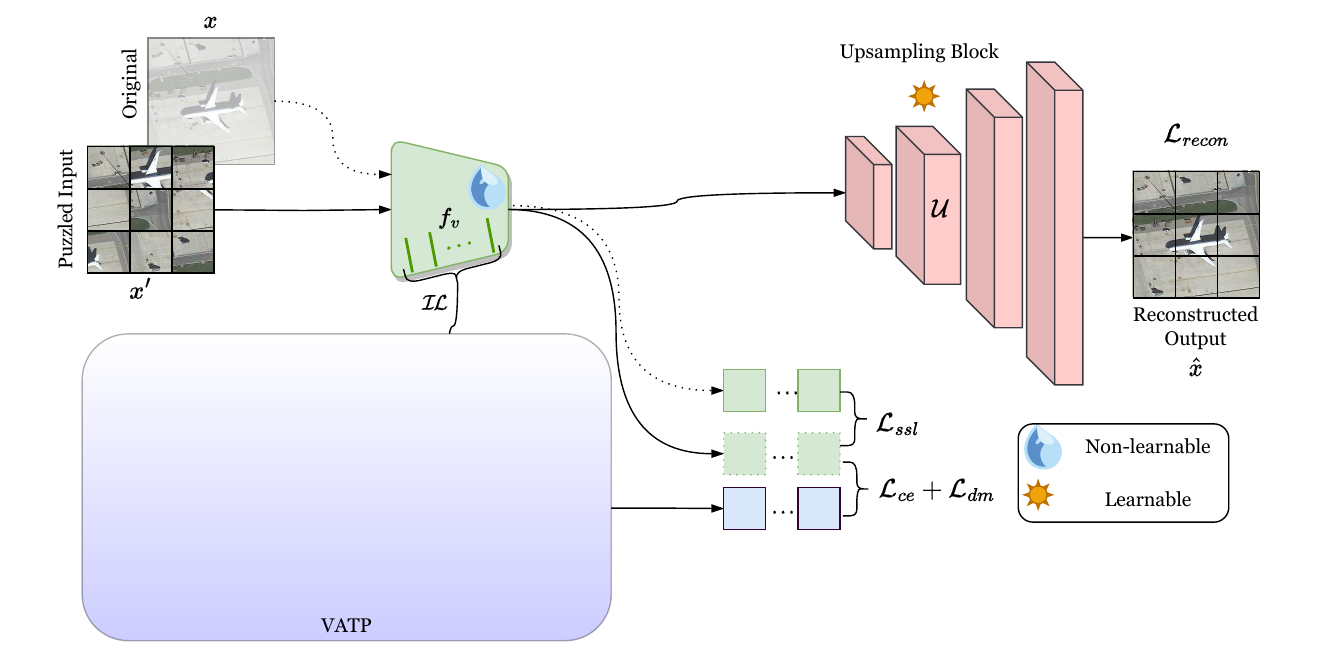}
    \caption{Our proposed \texttt{C-SAW} architecture utilizes $f_v$ and $f_t$ as the visual and text encoders, respectively, from CLIP's frozen backbone. The visual attentive token generator $\mathcal{G}_{VAT}$ generates $\mathcal{M}$ visual attentive tokens $v{\mathcal{S}}$ using intermediate layers $\mathcal{IL}$ of the source domains $\mathcal{S}$. These visual attentive tokens, along with context and class tokens, create text embeddings using $f_t$, forming the visual attentive text prompting (VATP) approach. We utilize these final image-conditioned prompt embeddings for the classification task under the supervision of the cross-entropy loss ($\mathcal{L}_{ce}$). Additionally, the upsampling block $\mathcal{U}$, represented in \textcolor{candypink}{light pink}, is associated with the reconstruction loss $\mathcal{L}_{recon}$, self-supervised loss ($\mathcal{L}_{ssl}$), and diversity maxmization loss ($\mathcal{L}_{dm}$). Best viewed in colour.}
    \label{fig:enter-label}
\end{figure*}
\section{Problem Definition \& Methodology}
We define dataset $\mathcal{D}= \{\mathcal{D}_S \cup \mathcal{D}_T\}$, where $\mathcal{D}_S$ and $\mathcal{D}_T$ represent the source and target domains, respectively, with each domain containing input data $\mathcal{X}$ and corresponding labels $\mathcal{Y}$. The probability distributions $P(\mathcal{D}_S^i)$ can vary for each domain. During training, we utilize labels $\mathcal{Y}_S$ from the source domains $\mathcal{D}_S$, while during testing, our focus shifts to labels $\mathcal{Y}_T$ from a distinct target test domain $\mathcal{D}_T$. The probability distribution $\mathcal{P}(\mathcal{D}_T)$ differs from $\mathcal{P}(\mathcal{D}_S^i)$ $\forall$ i. For base-to-new class generalization, there is no overlap between the label sets $\mathcal{Y}_S$ and $\mathcal{Y}_T$. In single-source domain generalization, the label sets are identical, resulting in $\mathcal{Y}_S \cap \mathcal{Y}_T = \mathcal{Y}_S \cup \mathcal{Y}_T$. The cross-dataset domain generalization presents variable situations of overlapping or non-overlapping label sets, warranting further exploration. This analysis provides valuable insights into domain adaptation and the relationships between source and target domains in diverse generalization scenarios.

\subsection{The Proposed \texttt{C-SAW}:}
In our paper, we introduce \texttt{C-SAW}, a novel approach that enhances text token learning and leverages the effectiveness of CLIP's visual backbone with jumbled input images ($x^{\prime}$) as presented in Figure \ref{fig:enter-label}. The pre-trained CLIP's vision encoder $f_v$ extracts features from both the original input image ($x$) and the jumbled input image ($x^{\prime}$), enabling contrastive generalization of images with text embeddings from the pre-trained CLIP's text encoder $f_t$. To achieve domain-agnostic style features for domain generalization, we utilize the mean $\mu$ of a batch of features from the intermediate layers ($\mathcal{IL}$) of CLIP's vision encoder. We observed that CLIP faces challenges in generalizing classification tasks when generating prompts from part-embeddings of $x^{\prime}$. To overcome this, our proposed \texttt{C-SAW} generates visual attentive tokens using $\mathcal{G_{VAT}}$. Additionally, we optimize \texttt{C-SAW} using the following losses; $\mathcal{L}_{ce}$ for classification, $\mathcal{L}_{dm}$ for contextualizing pre-trained CLIP's vision encoder for generalizing over part-embeddings of jumbled input images, and $\mathcal{L}_{ssl}$ for self-supervised learning to improve generalization. Furthermore, we use $\mathcal{L}_{recon}$ to reconstruct the original image from the jumbled input image, which helps in strengthening the latent space of $x'$. In the following subsections, we provide detailed explanations of our key contributions. 

\noindent\textbf{Self-supervised Block:} 
We propose an upsampling network ($\mathcal{U}$) to upscale the jigsaw or jumbled visual features from the vision encoder $f_v(x')$, aiming to reconstruct the original image $x$. Our upsampling network ($\mathcal{U}$) consists of four convolutional (CNN) layers, where the output dimensions of the first three CNN layers are reduced by half of the input dimensions using a kernel size of $7$, a stride of $3$, padding of $1$, and output padding of $2$. The channel dimension of the last CNN layer is set to match the input channel dimension of the RGB image. Finally, we reshape the output of the last layer through bilinear interpolation to match the shape of the original images, i.e., $\mathcal{U}(f_v(x'))=\hat{x} \in \mathbb{R} ^{3 \times 224 \times 224}$.


We further strengthen the contextualizing ability of CLIP's vision encoder $f_v$ part embeddings, using self-supervision between visual features $f_v(x')$ and $f_v(x)$. We define the reconstruction and self-supervised losses, i.e., $\mathcal{L}_{recon}$ and $\mathcal{L}_{ssl}$, in the subsequent paragraphs.

\vspace{0.3cm}
\noindent\textbf{Visual Attentive Text Prompting (VATP):} We derive the style representation for each domain by averaging the feature statistics $\mu$ of each batch $B$ from the respective domain. These statistics are obtained from the intermediate layers $\mathcal{IL}$ of the visual encoder of pre-trained CLIP. The style representation for the $i^{th}$ source domain is denoted as $\mu_{s_i}$. The generative visual attentive token $\mathcal{G_{VAT}}$ block takes the style information $\mu_s$ to generate $\mathcal{M}$ attentive tokens $v_{\mathcal{S_M}}$ conditioned on the input image $x$. Mathematically, this is denoted as $v_{\mathcal{S_M}} = \mathcal{G_{VAT}}(\mu(x'))$. The $\mathcal{G_{VAT}}$ block consists of $n$ linear layers, each for extracting style features from $n$ intermediate layers $f_v^{l}$, where $l\in\{1,\cdots,n\}$ and $f_v^l \in \mathbb{R}^{W \times H \times C}$ with $W$, $H$ and $C$ denote height, width and channel, respectively.

To generate the visual attentive tokens, we first create an attention mask $A(f_v^l(x')) = \texttt{sign}(\mathcal{G_{VAT}}(f_v^l(x')))$ by passing the obtained style features through a two-layer bottleneck structure, \texttt{Linear-ReLU-Linear-Sigmoid}. We then multiply and add this attention mask with the style features $f_v^l(x')$ in a residual manner to obtain the $\mathcal{M}$ visual attentive tokens $v_{\mathcal{S_M}}$, which is defined as
\begin{equation}
    v_{\mathcal{S_M}}(x') = [A(f_v^l(x')) \odot f_v^l(x')]+f_v^l(x')
\end{equation}
These visual attentive tokens $v_{\mathcal{S_M}}(x')$ are combined with textual tokens $c_{\mathcal{M}}$, i.e., $c^{\sim}_{\mathcal{M}}(x') = c_{\mathcal{M}} + v_{\mathcal{S_M}}(x')$. Finally, they pass through CLIP's pretrained text encoder $f_t$ to obtain the prompt token embeddings $prompt_{emb}(x')$, mathematically defined as
\begin{equation}
    prompt_{emb}(x') = f_t(c^{\sim}_{\mathcal{M}}(x')) \in\mathbb{R}^{K\times512}
\end{equation} 
Where $K$ denotes the number of classes. We follow the same process on the original input image $x$ to obtain $prompt_{emb}(x)$. Finally, we compute the average of both prompt embeddings: 
\begin{equation}
\begin{aligned}
   APE (x', x) = prompt_{emb_{avg}}(x', x)=&\\\dfrac{prompt_{emb}(x')+ prompt_{emb}(x)}{2}
\end{aligned}
\end{equation}
which allowing us to perform the classification contrastively with the visual feature of $x'$, i.e., $f_v(x')$.\\

\subsection{Losses and Network Optimization}
\noindent\textbf{Cross-entropy Loss:}
We classify the jumbled images ($x^{\prime}$) of the original images ($x$) in the contranstive manner with average text prompts generated with the respective conditioned input images i.e., $APE(x', x)$. The prediction probability for $x^{\prime}$ to belong to the label $y$ is denoted by,
\begin{equation}
p(y|x^{\prime}) = \frac{ \exp(<f_v(x^{\prime}), APE(x', x)>/\tau)}{\sum_{k=1}^{|\mathcal{Y}|}\ \exp(<f_v(x^{\prime}), APE(x', x)>/ \tau)}
\end{equation}
Here, $<\cdot>$ represents the cosine similarity, while $\tau$ stands for the temperature hyper-parameter. To compute the cross-entropy loss ($\mathbf{L_{ce}}$), we compare the prediction probabilities of each input image with their corresponding class labels as follows:
\begin{equation}
\label{eq:2}
    \mathcal{L}_{ce} = \underset{\mathcal{G_{VAT}}}{\arg\min} \underset{(x^{\prime},x,y) \in \mathcal{P}(\mathcal{D}_s)}{\mathbb{E}}  - \sum_{k=1}^{\mathcal{Y}_{S}} y_{k} log(p(y_k|x^{\prime}))
\end{equation}

\noindent\textbf{Self-supervised Loss:} We integrate \textsc{Barlow Twins} self-supervision \cite{barlowtwins} to align features from $f_v(x^{\prime})$ and $f_v(x)$, vision encoders for jumbled and original images, name as $\mathcal{L}_{ssl}$. This objective minimizes the discrepancy between cross-correlation and identity matrices, resulting in similar embeddings for distorted samples, removing redundancy. Our approach improves feature learning, leading to more coherent and efficient representations.

\vspace{0.2cm}
\noindent\textbf{Reconstruction Loss:} To achieve better peak signal-to-noise ratio between the jigsaw puzzle images ($x^{\prime}$) and reconstructed images ($\hat{x} = \mathcal{U}(f_v(x^{\prime}))$), we incorporate the $l_2$-norm between which is defined as,
\begin{equation}
    \label{eq:3}
    \mathcal{L}_{recon} = \underset{\mathcal{U}}{\text{argmin}} \underset{\mathcal{P}(\mathcal{D}_s)}{\mathbb{E}} || \hat{x} - x^{\prime} ||_2
\end{equation}
\noindent\textbf{Diversity Maximization Loss:} Additionally, we enforce a limitation on the similarity distribution between the visual features $f_v(x')$ of the target samples and prompt embeddings, by minimizing the entropy of prediction probabilities. It is defined as, 
\begin{equation}
\label{eq:4}
    \mathcal{L}_{dm} = \underset{p(y|x')} {\arg\min} \underset{\mathcal{P}(\mathcal{D}_s)}{\mathbb{E}} {min}([p(y_1|x');\cdots;p(y_{|\mathcal{Y}|}|x')])
\end{equation}

\vspace{0.1cm}
\noindent \textbf{Total Loss:} We optimize our proposed \texttt{C-SAW} with total loss, $\mathbf{L}_{total}$ is computed as:
\begin{equation}
\centering
    \label{eq:5}
    \begin{aligned}
    \mathcal{L}_{total} = \underset{ p(y|x'), \mathcal{U}, \mathcal{G_{VAT}}}{\arg\min} [\mathcal{L}_{ce} & + \alpha * (\mathcal{L}_{ssl} + \mathcal{L}_{recon}) \\
    & + (1 - \alpha) * \mathcal{L}_{dm}]
    \end{aligned}
\end{equation}
where $\alpha$ is a weight ratio factor associated for optimally balancing the ssl and diversity maximization losses.

\section{Experimental Results}
\label{experiments}
\noindent\textbf{Datasets:} For our experiments, we have used five remote sensing benchmark datasets i.e. PatternNet \cite{patternnet}, RSICD \cite{rsicd}, RESISC45 \cite{resisc45}, MLRSNet \cite{mlrsnet} and EuroSat \cite{helber2019eurosat}. The PatternNet dataset comprises $38$ classes, with each class consisting of $800$ images sized ($256 \times 256$) pixels. RSICD contains $30$ classes and a total of $10,000$ images, each with a size of $224 \times 224$ pixels. It's important to note that each class within RSICD has a varying number of images. The RESISC45 dataset consists of $45$ classes, with each class containing $700$ images sized ($256 \times 256$) pixels. MLRSNet, on the other hand, comprises 46 classes and a total of $109,161$ images, each sized ($256 \times 256$) pixels. The Eurosat consists of $10$ classes with total $27,000$ geo-referenced images. 

Additionally, we also work on generating learnable prompts within the single-source multi-target (SSMT) domain generalization setups, as mentioned in APPLeNet \cite{applenet}. The mentioned curated dataset consists of 16 overlapping classes that are common across all four datasets and suits for domain generalization tasks.\\

\noindent\textbf{Implementation Details:} We implemented our method in PyTorch on a 12GB Nvidia RTX 3090-Ti GPU card. Our proposed \texttt{C-SAW} is trained for 50 epochs using stochastic gradient descent (SGD) optimizer \cite{stochastic}. The initial learning rate is set to $2e^{-4}$ with a warm-up fixed rate of $1e^{-7}$ to prevent explosive gradients. Input images are rescaled to ($224\times224$) pixels and fed into CLIP's frozen encoder (ViT-B/16 \cite{vit}) for a latent dimension of $\mathbb{R}^{512}$. The model is trained with 16 samples per class and a batch size of 4. We experimentally choose the $\alpha$ values to be in between [0.5, 0.7], shown in Figure \ref{fig:alpha}. 

To initialize text prompts, we use \texttt{"a photo of a [CLS]"} embeddings, following previous literature \cite{coop, cocoop, applenet}, resulting in a context length of four. Three different seeds are used for evaluation, and the average \texttt{top-1} accuracy is reported. We experimentally select the parameters for optimizer, input preprocessing, and prompt initialization for optimal learning and convergence our proposed \texttt{C-SAW}.

\subsection{Comparison}
In this section, we conduct a thorough comparison between our novel approach, \texttt{C-SAW}, and existing methods in the realm of deep learning. We assess their performance across three different domain generalization (DG) tasks: \textbf{Base-to-Novel (B2N) Class Generalization}: This task involves training and testing the model on separate sets of classes without any overlap between them.
\textbf{Cross-Dataset (CD) Generalization}: Here, the model is trained on one dataset and then tested on new datasets that have variations in both their domains and labels. \textbf{Single Source Multi-Target (SSMT) Domain Generalization}: In this task, the model is trained on a specific source domain and then evaluated on multiple new domains, all within a closed-set scenario. To evaluate the effectiveness of our proposed \texttt{C-SAW}, we compare it against several established techniques, including zero-shot (ZS) CLIP \cite{clip}, ERM \cite{erm}, and DANN \cite{dann} for the SSMT task. Furthermore, we benchmark \texttt{C-SAW} against state-of-the-art (SOTA) prompt learning methods like CoOp \cite{coop}, CoCoOp \cite{cocoop}, CLIP-Adapter \cite{clip-adapter}, ProGrad \cite{prograd}, MaPLe \cite{maple}, APPLeNet \cite{applenet} and StyLIP \cite{stylip}, which serve as \textbf{baseline} methods for all the generalization tasks under consideration.\\


\noindent\textbf{B2N class generalization:} Table \ref{tab:B2N} presents the experimental results for Base-to-Novel (B2N) class generalization across five remote sensing (RS) datasets. The Table \ref{tab:B2N} includes the computation of the harmonic mean (HM), which represents the balance between the classification accuracies of the base and novel classes. For defining the source and target domains, we randomly and equally divide each dataset into two groups. We compare the performance of \texttt{C-SAW} with optimization-based methods that rely on referred context. Our proposed \texttt{C-SAW} method outperforms SOTA methods on the PatternNet, RSICD, RESISC45, MLRSNet and EuroSat datasets by margins of at least 3.5\%, 1.3\%, 2.0\%, 3.1\%, and 1.3\% respectively, when considering the harmonic mean of base and novel classes. Among the preferred methods, MaPLe shows significant performances over the non-RS methods and holds the third-best performance. APPLeNet achieves the second-best performance in generalizing the unseen classes across all RS datasets. However, when compared to CLIP's zero-shot approach, \texttt{C-SAW} demonstrates superior generalization scores, with a substantial margin of 37.33\% for seen classes and 10.18\% for unseen classes averaged across all the remote sensing datasets.

The impressive performance gains of \texttt{C-SAW} in the B2N class generalization tasks highlight its effectiveness in adapting to novel classes in remote sensing datasets. The results reinforce \texttt{C-SAW}'s capability to achieve robust domain generalization across diverse RS datasets and outperform existing optimization-based and zero-shot approaches. These findings position \texttt{C-SAW} as a competitive and reliable solution for class generalization in remote sensing domain adaptation tasks.
\vspace{0.3cm}
\begin{table*}[!ht]
\scriptsize{
    \centering   
    \caption{{Comparing \texttt{C-SAW} with SOTA methods on base-to-new class  generalization over existing methods on 5 different remote sensing datasets on $16$-shots with context length, $\mathcal{M}$=4. HM represents the harmonic mean.}
    \label{tab:B2N}}
    \hspace{0.2cm}
    \scalebox{1.4}{
    \begin{tabular}{ccclcccclcccclcc} 
     \multicolumn{4}{c}{(a) \textbf{Average over 5 datasets}}&\multicolumn{4}{c}{(b) PatternNet}&\multicolumn{4}{c}{(c) RSICD}\\\cmidrule(lr){1-4}\cmidrule(lr){5-8}\cmidrule(lr){9-12}
     
Method&\multicolumn{1}{c}{Base}&\multicolumn{1}{c|}{New}&\multicolumn{1}{c}{HM}&Method&\multicolumn{1}{c}{Base}&\multicolumn{1}{c|}{New}&\multicolumn{1}{c}{HM}&Method&\multicolumn{1}{c}{Base}&\multicolumn{1}{c|}{New}&\multicolumn{1}{c}{HM}\\\cmidrule(lr){1-4}\cmidrule(lr){5-8}\cmidrule(lr){9-12}

  CLIP \cite{clip}&\multicolumn{1}{c}{56.50}&\multicolumn{1}{c|}{60.04}&\multicolumn{1}{c}{58.22}&
 CLIP \cite{clip}&\multicolumn{1}{c}{63.67}&\multicolumn{1}{c|}{64.37}&\multicolumn{1}{c}{64.02 }&
 CLIP \cite{clip}&\multicolumn{1}{c}{54.61 }&\multicolumn{1}{c|}{ 55.33}&\multicolumn{1}{c}{ 54.97}\\

CoOp \cite{coop}&\multicolumn{1}{c}{89.65}&\multicolumn{1}{c|}{ 57.34}&\multicolumn{1}{c}{ 69.94}&
CoOp \cite{coop}&\multicolumn{1}{c}{91.62 }&\multicolumn{1}{c|}{62.23 }&\multicolumn{1}{c}{74.12 }&
CoOp \cite{coop}&\multicolumn{1}{c}{92.52}&\multicolumn{1}{c|}{56.08 }&\multicolumn{1}{c}{ 69.83}\\

CLIP-Adt \cite{clip-adapter}&\multicolumn{1}{c}{81.94}&\multicolumn{1}{c|}{ 59.05}&\multicolumn{1}{c}{ 68.64}&
CLIP-Adt \cite{clip-adapter}&\multicolumn{1}{c}{82.15 }&\multicolumn{1}{c|}{63.26 }&\multicolumn{1}{c}{71.48 }&
CLIP-Adt \cite{clip-adapter}&\multicolumn{1}{c}{78.93}&\multicolumn{1}{c|}{55.44 }&\multicolumn{1}{c}{ 65.13}\\
    
CoCoOp \cite{cocoop}&\multicolumn{1}{c}{87.71}&\multicolumn{1}{c|}{59.02 }&\multicolumn{1}{c}{ 70.56}&
CoCoOp \cite{cocoop}&\multicolumn{1}{c}{ 92.39}&\multicolumn{1}{c|}{63.34 }&\multicolumn{1}{c}{ 75.16}&
CoCoOp \cite{cocoop}&\multicolumn{1}{c}{ 93.18}&\multicolumn{1}{c|}{58.67}&\multicolumn{1}{c}{72.00 }\\

ProGrad \cite{prograd}&\multicolumn{1}{c}{87.55}&\multicolumn{1}{c|}{51.18 }&\multicolumn{1}{c}{ 64.60}&
ProGrad \cite{prograd}&\multicolumn{1}{c}{92.65 }&\multicolumn{1}{c|}{ 62.48}&\multicolumn{1}{c}{ 74.63}&
ProGrad \cite{prograd}&\multicolumn{1}{c}{ 93.44}&\multicolumn{1}{c|}{58.15 }&\multicolumn{1}{c}{71.69 }\\

MaPLe \cite{maple} &\multicolumn{1}{c}{90.85}&\multicolumn{1}{c|}{63.11}&\multicolumn{1}{c}{74.48} &
MaPLe \cite{maple} &\multicolumn{1}{c}{94.74}&\multicolumn{1}{c|}{66.12}&\multicolumn{1}{c}{77.88}&
MaPLe \cite{maple} &\multicolumn{1}{c}{94.91}&\multicolumn{1}{c|}{60.52}&\multicolumn{1}{c}{73.91}\\



APPLeNet \cite{applenet}& \multicolumn{1}{c}{90.95}&\multicolumn{1}{c|}{63.72}&\multicolumn{1}{c}{74.94}&
APPLeNet \cite{applenet}& \multicolumn{1}{c}{94.89}&\multicolumn{1}{c|}{ 65.57}&\multicolumn{1}{c}{ 77.55}&
APPLeNet \cite{applenet}& \multicolumn{1}{c}{\textbf{95.26}}&\multicolumn{1}{c|}{60.71 }&\multicolumn{1}{c}{74.16}\\

StyLIP \cite{stylip}& \multicolumn{1}{c}{91.25}&\multicolumn{1}{c|}{63.51}&\multicolumn{1}{c}{74.90}&
StyLIP \cite{stylip}& \multicolumn{1}{c}{95.13}&\multicolumn{1}{c|}{ 66.78}&\multicolumn{1}{c}{ 78.47}&
StyLIP \cite{stylip}& \multicolumn{1}{c}{94.98}&\multicolumn{1}{c|}{60.92 }&\multicolumn{1}{c}{74.23}\\

\cmidrule(lr){1-4}\cmidrule(lr){5-8}\cmidrule(lr){9-12}

\texttt{C-SAW}&\multicolumn{1}{c}{\textbf{92.90}}&\multicolumn{1}{c|}
{{\textbf{66.03}}}&\multicolumn{1}{c}{\textbf{77.20}}& 

\texttt{C-SAW}&\multicolumn{1}{c}{\textbf{96.03}}&\multicolumn{1}{c|}{\textbf{70.18}}&\multicolumn{1}{c}
{\textbf{81.09}}&

\texttt{C-SAW}&\multicolumn{1}{c}
{95.13}&\multicolumn{1}{c|}{\textbf{62.56}}&\multicolumn{1}{c}
{\textbf{75.48}}\\
\cmidrule(lr){1-4}\cmidrule(lr){5-8}\cmidrule(lr){9-12}
&&&&&&&&&&&\\
 \multicolumn{4}{c}{(d) RESISC45}&\multicolumn{4}{c}{(e) MLRSNet}&\multicolumn{4}{c}{(f) EuroSAT}&\multicolumn{4}{c}{}\\\cmidrule(lr){1-4}\cmidrule(lr){5-8}\cmidrule(lr){9-12}
 
 Method&\multicolumn{1}{c}{Base}&\multicolumn{1}{c|}{New}&
 \multicolumn{1}{c}{HM}&Method&\multicolumn{1}{c}{Base}&
 \multicolumn{1}{c|}{New}&\multicolumn{1}{c}{HM}&Method&\multicolumn{1}{c}{Base}&
 \multicolumn{1}{c|}{New}&\multicolumn{1}{c}{HM}\\\cmidrule(lr){1-4}\cmidrule(lr){5-8}\cmidrule(lr){9-12}

  CLIP \cite{clip}&\multicolumn{1}{c}{ 56.32}&
  \multicolumn{1}{c|}{55.38 }&\multicolumn{1}{c}{ 55.85}&
  CLIP \cite{clip}&\multicolumn{1}{c}{51.43 }&
  \multicolumn{1}{c|}{51.92}&\multicolumn{1}{c}{51.67 } & CLIP \cite{clip}&\multicolumn{1}{c}{56.48 }&
  \multicolumn{1}{c|}{64.05}&\multicolumn{1}{c}{60.03}\\

    CoOp \cite{coop}&\multicolumn{1}{c}{89.04 }&
    \multicolumn{1}{c|}{ 55.75}&\multicolumn{1}{c}{ 68.57}&
    CoOp \cite{coop}&\multicolumn{1}{c}{ 75.21}&
    \multicolumn{1}{c|}{53.64 }&\multicolumn{1}{c}{62.62}&CoOp \cite{coop}&\multicolumn{1}{c}{ 92.19}&
    \multicolumn{1}{c|}{57.74}&\multicolumn{1}{c}{68.69}\\

CLIP-Adt \cite{clip-adapter}&\multicolumn{1}{c}{81.67}&\multicolumn{1}{c|}{ 56.23}&\multicolumn{1}{c}{ 66.60}&
CLIP-Adt \cite{clip-adapter}&\multicolumn{1}{c}{71.64 }&\multicolumn{1}{c|}{53.19 }&\multicolumn{1}{c}{61.05 }&
CLIP-Adt \cite{clip-adapter}&\multicolumn{1}{c}{85.28}&\multicolumn{1}{c|}{61.07}&\multicolumn{1}{c}{71.17}\\

CoCoOp \cite{cocoop}&\multicolumn{1}{c}{ 89.78}&
\multicolumn{1}{c|}{ 57.18}&\multicolumn{1}{c}{ 69.86}&
CoCoOp \cite{cocoop}&\multicolumn{1}{c}{76.32 }&
\multicolumn{1}{c|}{ 52.75}&\multicolumn{1}{c}{62.38 }&
CoCoOp \cite{cocoop}&\multicolumn{1}{c}{87.49 }&
\multicolumn{1}{c|}{ 60.04}&\multicolumn{1}{c}{71.21}\\
ProGrad \cite{prograd}&\multicolumn{1}{c}{ 90.13}&\multicolumn{1}{c|}{57.89 }&\multicolumn{1}{c}{ 70.50}&
ProGrad \cite{prograd}&\multicolumn{1}{c}{75.96 }&\multicolumn{1}{c|}{ 52.23}&\multicolumn{1}{c}{ 61.90}&
ProGrad \cite{prograd}&\multicolumn{1}{c}{87.04 }&\multicolumn{1}{c|}{44.67 }&\multicolumn{1}{c}{59.04}\\

MaPLe \cite{maple} &\multicolumn{1}{c}{91.45}&\multicolumn{1}{c|}{60.82}&\multicolumn{1}{c}{73.05}&
MaPLe \cite{maple} &\multicolumn{1}{c}{79.06}&\multicolumn{1}{c|}{54.85}&\multicolumn{1}{c}{64.59}&MaPLe \cite{maple} &\multicolumn{1}{c}{94.07}&\multicolumn{1}{c|}{73.23}&\multicolumn{1}{c}{82.35}\\


APPLeNet \cite{applenet} & \multicolumn{1}{c}{91.24}&\multicolumn{1}{c|}{60.46 }&\multicolumn{1}{c}{72.73}&
APPLeNet \cite{applenet} & \multicolumn{1}{c}{78.53}&\multicolumn{1}{c|}{56.41}&\multicolumn{1}{c}{65.66}&
APPLeNet \cite{applenet} & \multicolumn{1}{c}{94.81}&\multicolumn{1}{c|}{75.46}&\multicolumn{1}{c}{84.04}\\

StyLIP \cite{stylip} & \multicolumn{1}{c}{90.87}&\multicolumn{1}{c|}{60.34 }&\multicolumn{1}{c}{72.52}&
StyLIP \cite{stylip} & \multicolumn{1}{c}{80.65}&\multicolumn{1}{c|}{55.47}&\multicolumn{1}{c}{65.73}&
StyLIP \cite{stylip} & \multicolumn{1}{c}{94.61}&\multicolumn{1}{c|}{74.06}&\multicolumn{1}{c}{83.08}\\

\cmidrule(lr){1-4}\cmidrule(lr){5-8}\cmidrule(lr){9-12}

\texttt{C-SAW}&\multicolumn{1}{c}{ \textbf{92.56}}&\multicolumn{1}{c|}
{\textbf{62.70} }&\multicolumn{1}{c}{ \textbf{74.76}}&
\texttt{C-SAW}&\multicolumn{1}{c}
{\textbf{85.41}}&\multicolumn{1}{c|}
{\textbf{57.52}}&\multicolumn{1}{c}
{\textbf{68.74} }&
\texttt{C-SAW}&\multicolumn{1}{c}
{\textbf{95.39}}&\multicolumn{1}{c|}
{\textbf{77.20}}&\multicolumn{1}{c}
{\textbf{85.33} }\\
\cmidrule(lr){1-4}\cmidrule(lr){5-8}\cmidrule(lr){9-12}
    \end{tabular}}}
\end{table*}


\noindent\textbf{CD generalization:} In the cross-dataset setup, we have evaluated \texttt{C-SAW} on the PatternNet dataset (source domain), and zero-shot inference results were reported for the remaining remote sensing (RS) datasets (target domains), as shown in Table \ref{tab:CDT}. Our method demonstrates remarkable improvements in both source and target classification performance. It surpasses zero-shot CLIP by substantial margins of 27.2\% in the source domain and 4.46\% on average across the target domains. Moreover, \texttt{C-SAW} outperforms SOTA methods by at least 2.55\% on average across the unseen target datasets. These results highlight the effectiveness of \texttt{C-SAW} in reducing the generalization gap between a single source and multiple target domains, even in the presence of domain and label shifts, within the CD generalization technique. The significant performance gains achieved by \texttt{C-SAW} in the CD setup underscore its ability to generalize effectively across diverse and unseen RS datasets. This demonstrates its potential for practical applications in remote sensing domain generalization tasks, where models must perform well on previously unseen target domains. The superior performance of \texttt{C-SAW} compared to zero-shot CLIP and other SOTA methods further solidifies its position as a promising approach for domain generalization and transfer learning in the remote sensing domain.
\vspace{0.3cm}
\begin{table}[!ht]
\centering
\scriptsize{
    \centering
    \caption{Comparing \texttt{C-SAW} with SOTA methods for cross-dataset generalization with PatternNet dataset as the source domain and remaining RS datasets as the target domains.}
    \scalebox{0.85}{
    \begin{tabular}{lcccccc} 
    \toprule
    &\multicolumn{1}{c}{\textbf{Source}}&\multicolumn{4}{c}{\textbf{Target}} \\
     
    \cmidrule(lr){2-2}\cmidrule(lr){3-7}
     
    \multirow{-2}{*}{\textbf{Method}}&\multicolumn{1}{c}{\textbf{PatternNet}}&\multicolumn{1}{c}{\textbf{RSICD}}&\multicolumn{1}{c}{\textbf{RESISC45}}
    &\multicolumn{1}{c}{\textbf{MLRSNet}}
    &\multicolumn{1}{c}{\textbf{EuroSAT}}&\multicolumn{1}{c}{\textbf{Average}}\\
    
    \midrule
    CLIP \cite{clip} & 61.72 & 43.25 & 48.56 & 45.13 & 43.24 & 45.65  \\

    CoOp \cite{coop} & 85.23 & 42.53  & 49.34 & 44.50 & 46.51 & 45.46  \\

    CLIP-Adapter \cite{clip-adapter} & 74.27& 42.57  & 49.07  & 44.17 & 44.75 & 45.27  \\

    CoCoOp \cite{cocoop} & 85.95 & 43.61 & 49.53 & 44.72 & 46.82 & 45.95\\

    ProGrad \cite{prograd} & 86.14 & 41.25 & 48.26 & 44.12 & 45.97 & 44.54 \\

    MaPLe \cite{maple} & 87.92 & 45.23 & 49.56 & 46.37 & 47.63 & 47.05 \\
 APPLeNet \cite{applenet} & 88.17 & 44.87 & \textbf{50.97} & 46.83 & 49.52 & 47.56 \\
 
 StyLIP \cite{stylip} & 88.01 & 46.12 & 49.89 & 46.94 & 49.79 & 48.19 \\

 \midrule
 
 \texttt{C-SAW} & \textbf{88.92} & \textbf{50.47} & 50.60 & \textbf{49.26} & \textbf{50.54} & \textbf{50.11} \\\bottomrule
    \end{tabular}}\label{tab:CDT}}
\end{table}

\noindent\textbf{SSMT domain generalization:} In the Single-Source Multi-Target (SSMT) domain generalization setup, which differs from the previously discussed cross-dataset transfer (CDT) setting with shared classes across domains, we select the PatternNetv2 dataset as the target domain, following the approach of APPLeNet \cite{applenet}. The results presented in Table \ref{tab:ssmt} demonstrate the remarkable performance of \texttt{C-SAW}. Our proposed method has outperformed all the SOTA methods, showcasing a substantial lead of at least 4.67\% on RSICDv2, 5.35\% on RESISC45v2, and 4.23\% on the MLRSNetv2 datasets. 
Overall, \texttt{C-SAW} exhibited notably superior performance, achieving a $4.94\%$ improvement over the average performance across the target domains. The performance superiority of \texttt{C-SAW} in the SSMT setup further validates its effectiveness in addressing the domain generalization challenges in remote sensing images. These results underscore the potential of our approach in real-world applications, where adapting to diverse and unseen target domains is crucial. The robust performance gains achieved by \texttt{C-SAW} across multiple target domains highlight its promise as a practical and effective solution for remote sensing domain generalization tasks.

\begin{table}[!ht]
\centering
\scriptsize{
    \centering
    \caption{Comparing \texttt{C-SAW} with SOTA methods for single-source multi-target domain generalization on the benchmark RS datasets.}
    \scalebox{0.9}{
    \begin{tabular}{lccccc} 
    \toprule
    &\multicolumn{1}{c}{\textbf{Source}}&\multicolumn{4}{c}{\textbf{Target}} \\
     
    \cmidrule(lr){2-2}\cmidrule(lr){3-6}
     
      \multirow{-2}{*}{\textbf{Method}}&\multicolumn{1}{c}{\textbf{PatternNetv2}}&\multicolumn{1}{c}{\textbf{RSICDv2}}&\multicolumn{1}{c}{\textbf{RESISC45v2}}
    &\multicolumn{1}{c}{\textbf{MLRSNetv2}}&\multicolumn{1}{c}{\textbf{Average}}\\
    
    \midrule
    ERM \cite{erm} & 73.69  & 61.40  & 61.59 & 61.13 & 61.37  \\
    CLIP \cite{clip} & 78.04 & 72.15 & 75.42 & 67.78 & 71.78 \\
    
    DANN \cite{dann} & 93.56 & 75.49  & 76.18  & 70.53 & 74.07  \\

    CoOp \cite{coop} & 94.25 & 76.50  & 77.87& 70.97 & 75.11 \\

    CLIP-Adapter \cite{clip-adapter} & 92.36 & 79.17 & 79.76 & 71.04 & 76.66 \\

    CoCoOp \cite{cocoop} & 94.41 & 79.33 & 80.43 & 71.67 & 77.14 \\

    ProGrad \cite{prograd} & 95.18 & 77.46 & 80.65 & 72.29 & 76.80 \\

    MaPLe \cite{maple} & 96.52 & 80.45 & 83.37 & 76.15 & 79.99 \\

    
   APPLeNet \cite{applenet} & 96.63 & 81.03 & 82.23 & 74.03 & 79.10 \\

   StyLIP \cite{stylip} & 96.85 & 80.67 & 84.56 & 75.66 & 80.30 \\

   \midrule
   
   \texttt{C-SAW} & \textbf{97.91} & \textbf{85.70} & \textbf{88.72} & \textbf{80.38} & \textbf{84.93} \\  \bottomrule
    \end{tabular}}
\label{tab:ssmt}}
\end{table}

\subsection{Ablation Studies}

\begin{figure*}
    \centering
    \includegraphics[width = \textwidth]{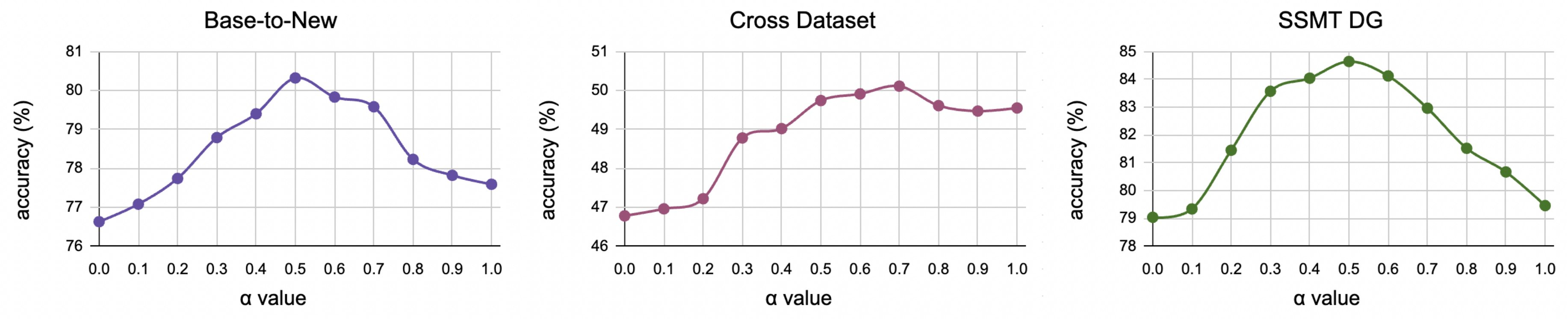}
    \vspace{-0.6cm}
    \caption{Ablation of weight ratio factor ($\alpha$) in all generalization setup.}
    
    \label{fig:alpha}
    
\end{figure*}
\noindent\textbf{Ablation study on weight ratio factor ($\alpha$):} In this section, we have investigated the impact of the weight ratio factor ($\alpha$) on the performance of our model, \texttt{C-SAW}. $\alpha$ determines the balance between self-supervision and supervision tasks that \texttt{C-SAW} should carry to optimize its performance, as defined by Eq. \ref{eq:5}, where $\alpha$ and $(1-\alpha)$ represent the ratios in which the two tasks are involved, respectively. Figure \ref{fig:alpha} illustrates the results of varying the weight ratio factor $\alpha$. We have observed that the best performance is achieved when $\alpha = 0.5$, particularly in the B2N and SSMT setups. For the CD setup, the optimal weight ratio factor is found to be $\alpha = 0.7$. These findings indicate that striking an appropriate balance between self-supervision and supervision tasks is critical to maximizing the performance of \texttt{C-SAW} across different experimental setups. By fine-tuning the weight ratio factor, we ensure that \texttt{C-SAW} effectively leverages both forms of training signals, leading to superior performance in domain generalization tasks.\\

\noindent\textbf{Evaluation of \texttt{C-SAW}'s sensitivity to the number of shots:} The performance of our proposed \texttt{C-SAW} is assessed by varying the number of shots from 1 to all for the B2N class generalization task. A comparison is made against SOTA prompting techniques, as presented in Table \ref{tab:shots}. For this evaluation, we utilize a context length of 4, position the class token at the end, employ ViT-B/16 \cite{vit} as the visual feature backbone, and utilize a unified context vector. Since CLIP operates in a zero-shot manner, it is excluded from this comparison, and we focus solely on few-shot-based prompting methods while showcasing results on the average of all RS datasets. Our results demonstrate that \texttt{C-SAW} surpasses the performance of benchmark prompt learning-based methods by at least $1.9\%$, $2.2\%$, $2.3\%$, $2.4\%$ and $2.1\%$ for 1, 4, 8, 16 shots and all images, respectively. \\

\begin{table}[!ht]

\centering
\scriptsize{
    \centering
    \caption{Comparing \texttt{C-SAW} with the SOTA methods on varying the number of shots for the B2N class generalization (considered harmonic mean HM) task with average of all RS datasets.}
    \scalebox{1.2}{
    \begin{tabular}{lccccc} 
    \toprule
     
    \multirow{-1}{*}{\textbf{Method}}&\multicolumn{1}{c}{\textbf{1-shot}}&\multicolumn{1}{c}{\textbf{4-shot}}&\multicolumn{1}{c}{\textbf{8-shot}}
    &\multicolumn{1}{c}{\textbf{16-shot}}&\multicolumn{1}{c}{\textbf{All}}\\
    
    \midrule
    CoOp \cite{coop} & 65.31 & 66.85 & 67.52 & 69.94 & 67.68  \\

    CLIP-Adapter \cite{clip-adapter} & 63.40 & 64.58 & 67.11 & 68.64 & 66.57  \\

    CoCoOp \cite{cocoop} & 66.82 & 68.31 & 69.03 & 70.56 & 68.85 \\

    ProGrad \cite{prograd} & 59.33 & 61.16 & 63.44 & 64.60 & 60.86  \\

    MaPLe\cite{maple} & 73.42 & 74.28 & 76.63 & 77.49 & 76.72\\

    APPLeNet\cite{applenet} & 75.82 & 76.23 & 77.69 & 78.43 & 78.51\\
  
    \midrule
   
   \texttt{C-SAW} & \textbf{77.14} &\textbf{77.83} & \textbf{79.42} & \textbf{80.32} & \textbf{79.95} \\\bottomrule
    \end{tabular}}\label{tab:shots}}
\end{table}

\noindent\textbf{Ablation study on loss terms:}
We have performed several experiments with our proposed model, \texttt{C-SAW}, employing different loss terms, as outlined in Table \ref{tab:loss}. One of these loss terms, $SSL$ defines the self-supervision importance of the model. It combines the self-supervised loss ($\mathcal{L}_{ssl}$) and reconstruction loss ($\mathcal{L}_{recon}$). They are commonly used to minimize the gap between self-supervised views and the original images. By taking only these losses with $\mathcal{L}_{ce}$, decreases the model performance by $2.73\%$, $0.56\%$ and $5.18\%$ in B2N, CD and SSMT set up respectively. For that reason, we have considered a $Non$-$SSL$ loss $\mathcal{L}_{dm}$. However, considering $\mathcal{L}_{dm}$ only and removing $SSL$ losses lead to a performance decrease of $3.69\%$, $3.33\%$ and $5.60\%$ in B2N, CD and SSMT set up respectively. From Eq. \ref{eq:5}, here we have considered when $\alpha=1.0$, only $SSL$ loss works, when $\alpha=0$, only $\mathcal{L}_{dm}$ contributes and when $\alpha=0.5$, the best output comes for B2N and SSMT DG tasks as discussed above in details on the importance of $\alpha$ factor.
\begin{table}[!ht]
\centering
\scriptsize{
    \centering
    \caption{Ablation study of \texttt{C-SAW} with different settings of losses in all generalization setup over average of all RS datasets. Here, we define $SSL$  $(\mathcal{L}_{ssl}+\mathcal{L}_{recon})$.}
    \scalebox{1.4}{
    \begin{tabular}{lccc} 
    \toprule
     
    \multirow{-1}{*}{\textbf{Loss}}&\multicolumn{1}{c}{\textbf{B2N}}&\multicolumn{1}{c}{\textbf{CD}}&\multicolumn{1}{c}{\textbf{SSMT}}\\
    
    \midrule
    $\mathcal{L}_{ce}$ & 72.93 & 45.82 & 76.39    \\

    $\mathcal{L}_{ce} + SSL$ & 77.59 & 49.55 & 79.46 \\

    $\mathcal{L}_{ce} + \mathcal{L}_{dm}$ & 76.63 & 46.78 & 79.04 \\

    $\mathcal{L}_{ce} + SSL + \mathcal{L}_{dm}$  & \textbf{80.32} & \textbf{50.11} & \textbf{84.64}  \\
    
    \bottomrule
    \end{tabular}}\label{tab:loss}}
\end{table}

\noindent\textbf{Ablation study on context lengths:}
The context length four \texttt{"a photo of a [CLS]"} is common and experimentally found to be optimal in SOTA prompt-learning methods \cite{maple,cocoop, prograd}. As shown in Table \ref{tab:context}, we have experimented and found that the context length of $4$ provides best performance in comparison with $M=8, 12$ and $16$. 

\begin{table}[!ht]
\centering
\scriptsize{
    \centering
    \caption{Comparing \texttt{C-SAW} with the SOTA methods on varying context length for the B2N class generalization.}
    \scalebox{1.4}{
    \begin{tabular}{lcccc} 
    \toprule
     
 {\textbf{Context Length ($M$)}}&\multicolumn{1}{c}{\textbf{4}}&\multicolumn{1}{c}{\textbf{8}}
    &\multicolumn{1}{c}{\textbf{12}}&\multicolumn{1}{c}{\textbf{16}}\\
\midrule
   \texttt{C-SAW}  &\textbf{80.32} & 79.55 & 79.02 & 77.39 \\\bottomrule
    \end{tabular}}\label{tab:context}}
      \vspace{-0.32cm}
\end{table}

\begin{table}[!ht]
\centering
\scriptsize{
    \centering
    \caption{Comparing \texttt{C-SAW} with different feature extractors for single-source multi-target domain generalization on the benchmark RS datasets.}
    \scalebox{1.05}{
    \begin{tabular}{lccccc} 
    \toprule
    &\multicolumn{1}{c}{\textbf{Source}}&\multicolumn{4}{c}{\textbf{Target}} \\
     
    \cmidrule(lr){2-2}\cmidrule(lr){3-6}
     {\textbf{Method}}&\multicolumn{1}{c}{\textbf{PatternNetv2}}&\multicolumn{1}{c}{\textbf{RSICDv2}}&\multicolumn{1}{c}{\textbf{RESISC45v2}}
    &\multicolumn{1}{c}{\textbf{MLRSNetv2}}&\multicolumn{1}{c}{\textbf{Average}}\\
    
    \midrule
    RN50\cite{resnet50} & 65.12  & 54.30  & 52.77 & 53.45 & 53.51  \\
    DINO\cite{caron2021emerging} & 80.55 & 74.10 & 71.92 & 75.39 & 73.80 \\
   
   \texttt{C-SAW} & \textbf{97.91} & \textbf{85.70} & \textbf{88.72} & \textbf{80.38} & \textbf{84.93} \\  \bottomrule
    \end{tabular}}
\label{tab:ssmt_feature}}
    \vspace{-0.2cm}
\end{table}

\noindent\textbf{Ablation study on feature extractors:}
Since CLIP is the most popular SOTA for few-shot or zero-shot tasks in computer vision, we use CLIP as the frozen network for feature extractor and perform prompt tuning for various downstream tasks. We also compare our proposed \texttt{C-SAW} with simple vision extractors like \textit{ResNet-50 (RN50) \cite{resnet50} and DINO \cite{caron2021emerging}} on SSMT DG task, and we can clearly observe that prompt learning conditioned to image features outperforms pre-trained feature extractors at least by $11 \%$ as mentioned in Table \ref{tab:ssmt_feature}.

\section{Conclusions}
We present \texttt{C-SAW}, a framework designed to enhance the multi-domain generalization capability of CLIP-derived features by introducing two key improvements on top of the frozen CLIP model. Firstly, we propose a jigsaw-based self-supervised objective to supplement CLIP's vision encoder. This addition injects the importance of part-aware visual feature learning, effectively addressing a limitation present in the baseline CLIP model. Secondly, we introduce a novel prompt learning approach within \texttt{C-SAW}, strategically integrating visual content and style primitives into the prompts. This integration enables the model to achieve better generalization to previously unseen domains and classes. Through these innovative modifications, \texttt{C-SAW} demonstrates impressive performance in dealing with challenging optical remote sensing images, by achieving better generalization across diverse domains and classes. In the future, we aim to further enhance the model's capabilities by incorporating outlier identification ability, unlocking even more potential for anomaly detection and handling challenging scenarios.

\bibliographystyle{ACM-Reference-Format}
\bibliography{cite}

\appendix




\end{document}